\pdfoutput=1

\documentclass[11pt]{article}

\usepackage[]{EMNLP2022}

\usepackage{times}
\usepackage{latexsym}

\usepackage[T1]{fontenc}

\usepackage[utf8]{inputenc}

\usepackage{microtype}

\usepackage{amsmath}
\usepackage{amssymb}
\usepackage{framed} 
\usepackage{subcaption}  
\usepackage{tabularx}  
\usepackage{booktabs}  
\usepackage{multirow}  
\usepackage{makecell}
\makeatletter
\def\new@fontshape{}
\makeatother
\usepackage{siunitx} 

\usepackage{xspace}  

\usepackage{tikz}  
\usetikzlibrary{shapes.geometric, arrows}
\usetikzlibrary{calc, positioning}  
\usepackage{forest}  
\useforestlibrary{linguistics}
\forestapplylibrarydefaults{linguistics}

\usepackage{stmaryrd} 

\usepackage[compress,sort,capitalise]{cleveref}  
\usepackage{hyperref}

\usepackage[inline]{enumitem}  

\usepackage{gb4e}  
\noautomath

\newcommand{\lform}[1]{\texttt{#1}}

\crefname{xnumi}{}{}
\creflabelformat{xnumi}{(#2#1#3)}
\crefrangeformat{xnumi}{(#3#1#4)--(#5#2#6)}

\tikzstyle{nnode} = [ellipse, text centered, draw=black, inner sep=2pt] 
\tikzstyle{dnode} = [rectangle, rounded corners, text centered, draw=black] 
\tikzstyle{arrow} = [thick,->,>=stealth]

\definecolor{colorwant}{HTML}{ABBEF1} 
\definecolor{colorgo}{HTML}{DE8BB3} 
\definecolor{colorthe}{HTML}{F1AD83} 
\definecolor{colorboy}{HTML}{FFE19D} 

\pgfdeclarelayer{background}
\pgfdeclarelayer{main}
\pgfsetlayers{background,main}
\newcommand*{\minWidth}{25}  
\newcommand*{\maxValue}{100}
\newcommand{\makeBar}[3]{%
  \tikz[baseline]{
    \node[anchor=base,text width=\minWidth,align=#2,inner sep=0pt,inner xsep=4pt,outer sep=0pt] (n) {\strut{\hfill#1}};  
    \begin{pgfonlayer}{background}
        {
       \edef\color{#3}
       \pgfmathparse{abs(#1/\maxValue)}
       \edef\contents{{\pgfmathresult}}
       \fill[font=\boldmath,color=\color] (n.north west) rectangle ($(n.south west)!{{\contents}}!(n.south east)$);  
       \fill[font=\boldmath,color=cyan!10] ($(n.north west)!{{\contents}}!(n.north east)$) rectangle (n.south east);  
       }
    \end{pgfonlayer}
  }
}
\newcommand{\asbar}[1]{\makeBar{#1}{center}{cyan!25}}

\newcommand{\genclass}[1]{\textsc{#1}\xspace} %
\newcommand{\lexg}{\genclass{Lex}}

\newcommand{\structg}{\genclass{Struct}}

\usepackage{color}
\usepackage{bm}
\definecolor{orange}{rgb}{1,0.5,0}
\definecolor{mdgreen}{rgb}{0.05,0.6,0.05}
\definecolor{mdblue}{rgb}{0,0,0.7}
\definecolor{dkblue}{rgb}{0,0,0.5}
\definecolor{dkgray}{rgb}{0.3,0.3,0.3}
\definecolor{slate}{rgb}{0.25,0.25,0.4}
\definecolor{gray}{rgb}{0.5,0.5,0.5}
\definecolor{ltgray}{rgb}{0.7,0.7,0.7}
\definecolor{purple}{rgb}{0.7,0,1.0}
\definecolor{lavender}{rgb}{0.65,0.55,1.0}
\definecolor{brown}{rgb}{0.6,0.2,0.2}

\newcommand{\code}[1]{}

\newcommand{\linf}{structure-aware}
\newcommand{\Linf}{Structure-aware}

\newcommand{\cogsqa}{QA-COGS}
\newcommand{\cogsqabase}{\cogsqa -base}
\newcommand{\cogsqacc}{\cogsqa -disamb}

\newcommand{\syncogs}{Syntax-COGS}
\newcommand{\poscogs}{POS-COGS}

\renewcommand{\paragraph}[1]{\textbf{#1}}
\title{Structural generalization is hard for sequence-to-sequence models}

\author{Yuekun Yao \and Alexander Koller \\
  Department of Language Science and Technology\\
  Saarland Informatics Campus\\
  Saarland University, Saarbrücken, Germany \\
  \texttt{\{ykyao, koller\}@coli.uni-saarland.de} \\}

\begin{document}
\maketitle
\begin{abstract}
Sequence-to-sequence (seq2seq) models have been successful across many NLP tasks,
including ones that require predicting linguistic structure. However, recent work on compositional generalization
has shown that seq2seq models achieve very low accuracy in generalizing to linguistic structures
that were not seen in training.
We present new evidence that this is a general limitation of seq2seq models that is present
not just in semantic parsing, but also in syntactic parsing and in text-to-text tasks, and that this
limitation can often be overcome by neurosymbolic models that have linguistic knowledge built in.
We further report on some experiments that give initial answers on the reasons for these limitations.
\end{abstract}


\section{Introduction} \label{sec:intro}

Humans are able to understand and produce linguistic structures they have never observed before \citep{chomsky-1957-syntactic,fodor-pylyshyn-1988-connectionism,fodor-lepore-2002-compositionality}. From limited, finite observations, they generalize at an early age to an infinite variety of novel structures using recursion. They can also assign meaning to these, using the Principle of Compositionality. 
This ability to generalize to unseen structures is important for NLP systems in low-resource settings, such as underresourced languages or projects with a limited annotation budget, where a user can easily use structures that had no annotations in training.

Over the past few years, large pretrained sequence-to-sequence (seq2seq) models, such as BART \citep{lewis-etal-2020-bart-acl} and T5 \citep{raffel-etal-2020-t5}, have brought tremendous progress to many NLP tasks. This includes linguistically complex tasks such as broad-coverage semantic parsing, where e.g.\ a lightly modified BART set a new state of the art on AMR parsing \citep{bevilacqua-etal-2021-one}. However, there have been some concerns that seq2seq models may have difficulties with \emph{compositional generalization}, a class of tasks in semantic parsing where the training data is structurally impoverished in comparison to the test data \citep{lake-baroni-2018-generalization,keysers-etal-2020-measuring}. We focus on the COGS dataset of \citet{kim-linzen-2020-cogs}
because some of its generalization types specifically target \emph{structural generalization}, i.e.\ the ability to generalize to unseen structures.

In this paper, we make two contributions. First, we offer evidence that structural generalization is systematically hard for seq2seq models. On the semantic parsing task of COGS, seq2seq models don't fail on compositional generalization as a whole, but specifically on the three COGS generalization types that require generalizing to unseen linguistic structures, achieving accuracies below 10\%. This is true both for BART and T5 and for seq2seq models that were specifically developed for COGS. What's more, BART and T5 fail similarly on syntax and even POS tagging variants of COGS (introduced in this paper), indicating that they do not only struggle with \emph{compositional} generalization in semantics, but with \emph{structural} generalization more generally. \Linf\ models, such as the compositional semantic parsers of \citet{liu-etal-2021-learning} and \citet{weissenhorn22starsem} and the Neural Berkeley Parser \citep{kitaev-klein-2018-constituency}, achieve perfect accuracy on these tasks.

\begin{figure*}[htb!]
    \small
    \centering
    \setlength{\tabcolsep}{2pt}
    \begin{tabularx}{\linewidth}{p{1.7cm}XX} \toprule
         & \textbf{Training} & \textbf{Generalization} \\ \midrule
         (a) \lexg
         \newline {\tiny subj\_to\_obj}
         \newline {\tiny (common noun)}
         & {A \underline{hedgehog} ate the cake. \newline
         \lform{*cake($x_4$); \underline{hedgehog($x_1$)} $\land$ eat.agent($x_2, \underline{x_1}$) $\land$eat.theme($x_2, x_4$)}} 
         & {The baby liked the \underline{hedgehog}. \newline
         \lform{*baby($x_1$); *\underline{hedgehog($x_4$)}; like.agent($x_2, x_1$) $\land$ like.theme($x_2, \underline{x_4}$)} }\\ 
         \midrule
         (b) \structg 
         \newline {\tiny PP recursion}
         & Ava saw a ball in a bowl on the table. \newline
         \lform{*table($x_9$); see.agent($x_1,$ Ava) $\land$ see.theme($x_1, x_3$) $\land$ ball($x_3$) $\land$ ball.nmod.in($x_3, x_6$) $\land$ bowl($x_6$) $\land$ bowl.nmod.on($x_6,x_9$)}
         &  Ava saw a ball in a bowl on the table \underline{on the floor}. \newline
         \lform{*table($x_9$); \underline{*floor($x_{12}$);} see.agent($x_1,$ Ava) $\land$ see.theme($x_1, x_3$) $\land$ ball($x_3$) $\land$ ball.nmod.in($x_3, x_6$) $\land$ bowl($x_6$) $\land$ bowl.nmod.on($x_6,x_9$) \underline{$\land$ table.nmod.on($x_9,x_{12}$)}} \\ \cmidrule(l){2-3}
         (c) \structg 
         \newline {\tiny obj\_to\_subj PP} 
         & Noah ate \underline{the cake on the plate}. \newline
         \lform{*cake($x_3$); *plate($x_6$); \newline eat.agent($x_1,$ Noah) $\land$ eat.theme($x_1, x_3$) $\land$ cake.nmod.on($x_3,x_6$)} 
         & \underline{The cake on the table} burned. \newline
         \lform{*cake($x_1$); *table($x_4$); cake.nmod.on($x_1,x_4$) $\land$ burn.theme($x_3, x_1$)}\\ 
         \bottomrule
    \end{tabularx}
    \caption{
        Some examples from the COGS dataset. \lexg represents lexical generalization and \structg denotes structural generalization.
    }\label{fig:cogssamples}
\end{figure*}

Second, we conduct a series of experiments to investigate what makes structural generalization so hard for seq2seq models. It is not because the encoder loses structurally relevant information: One can train a probe to predict COGS syntax from BART encodings, in line with earlier work \citep{hewitt-manning-2019-structural,tenney-etal-2019-bert}; but the decoder does not learn to use it for structural generalization. We find further that the decoder does not even learn to generalize semantically when the input is enriched with syntactic structure. Finally, it is not merely because the COGS tasks require the mapping of language into symbolic representations. We introduce a new text-to-text variant of COGS called \emph{\cogsqa}, where  questions about COGS sentences must be answered in English. We find that T5 performs well on structural generalization with the original COGS sentences, but all models still struggle with a harder text-to-text task involving structural disambiguation. 

The code\footnote{\url{https://github.com/coli-saar/Seq2seq-on-COGS}} and datasets\footnote{\url{https://github.com/coli-saar/Syntax-COGS}} are available online.

\section{Related work}\label{sec:related_work}


The recent interest in compositional generalization has raised concerns about limitations of seq2seq models. For instance, the SCAN dataset \citep{lake-baroni-2018-generalization} requires a model to translate natural-language instructions into symbolic action sequences; it has multiple splits in which the test data contains new combinations of commands or instructions that are systematically longer than in training. The PCFG dataset \citep{hupkes2020compositionality} builds upon SCAN and adds instructions with recursive structure. The CFQ dataset \citep{keysers-etal-2020-measuring} maps questions to SPARQL queries, and splits the data according to a measure of compositional complexity (MCD). In all of these papers, simple seq2seq models based on LSTMs and transformers were shown to perform poorly when the test data was more complex than the training data.

Since then, followup research has shown that both generic transformer-based models \citep{ontanon-etal-2022-making,csordas-etal-2021-devil}, general-purpose pretrained models \citep{furrer-etal-2020-compositional}, and seq2seq models that are specialized for the task can achieve higher accuracies than the ones reported in the papers introducing the datasets. Nonetheless, there is a sense that despite the best efforts of the community, pure seq2seq models are hitting a ceiling on compositional generalization tasks. 

In this paper, we shed some light on the issue by (a) clarifying that seq2seq models do not struggle with compositional generalization per se, but with \emph{structural} generalization, and (b) demonstrating that this type of generalization remains hard for seq2seq models even after heavy pretraining. This is in contrast to most previous research, which has avoided pretraining and focused on length or MCD as the primary source of difficulty. Our data includes instances where the structure, but not the length differs between training and testing, and therefore allows us to differentiate between the two. The importance of structure to compositional generalization is also recognized by \citet{bogin-2022-local-structure}.

The difficulty of structural generalization for neural models has also been studied in more targeted ways. For instance, \citet{yu-etal-2019-learning} show empirically that LSTM-based seq2seq models cannot learn to close the brackets of Dyck languages, and \citet{hahn-2020-theoretical} proves that transformers cannot learn to distinguish well-bracketed Dyck expressions. 
\citet{DBLP:journals/tacl/McCoyFL20} find empirically that seq2seq models struggle to learn the structural operations necessary to rewrite declarative English sentences into questions, whereas tree-based models work better.

\section{Structural generalization in COGS}\label{sec:background}

\begin{figure}[htbp]
  \centering
  \begin{subfigure}[b]{\columnwidth}
    \centering
    \hbox{\tiny

\begin{forest} for tree={l=0, l sep=4, inner sep=1}
[S, s sep=1pt, 
  [NP [Noah]]
  [VP, s sep=10pt, 
    [V [ate]]
    [NP,tikz={\node [draw=black,fill=cyan!30, fill opacity=0.2,inner sep=0,fit to=tree]{};}
      [NP [the cake,roof]]
      [PP [on the plate,roof]]
    ]
  ]
]
\end{forest}
\hspace{-5mm}
$\implies$

\begin{forest} for tree={l=0, l sep=4, inner sep=1}
[S, s sep=1pt,  
  [NP,tikz={\node [draw=black,fill=cyan!30, fill opacity=0.2,inner sep=0,fit to=tree]{};}
      [NP [the cake,roof]]
      [PP [on the table,roof]]
  ]
  [VP
    [V [burned]]
  ]
]
\end{forest}
    }
    \caption{object PP to subject PP}\label{fig:structural-generalization:toSubjPP}
  \end{subfigure}
  
  \begin{subfigure}[b]{\columnwidth}
      \centering
      \hbox{\tiny

\begin{forest} for tree={l=0, l sep=3, s sep=6pt, inner sep=1}
[S 
  [NP [Ava]]
  [VP
    [V [saw]]
    [NP
      [NP [a ball,roof]]
      [PP,tikz={\node [draw=black,fill=cyan!30, fill opacity=0.2,inner sep=0,fit=()(!1)(!2)]{};}
       [in]
       [NP
         [NP [a bowl,roof]]
         [PP,tikz={\node [draw=black,fill=cyan!30, fill opacity=0.2,inner sep=0,fit=()(!1)(!2)]{};}
           [on]
           [NP [the table,roof]]
         ]
       ]
      ]
    ]
  ]
]
\end{forest}
\hspace{-1.5cm}
$\implies$

\begin{forest} for tree={l=0, l sep=3, s sep=6pt, inner sep=1}
[S 
  [NP [Ava]]
  [VP
    [V [saw]]
    [NP
      [NP [a ball,roof]]
      [PP,tikz={\node [draw=black,fill=cyan!30, fill opacity=0.2,inner sep=0,fit=()(!1)(!2)]{};}
       [in]
       [NP
         [NP [a bowl,roof]]
         [PP,tikz={\node [draw=black,fill=cyan!30, fill opacity=0.2,inner sep=0,fit=()(!1)(!2)]{};}
           [on]
           [NP 
             [NP [the table,roof]]
             [PP,tikz={\node [draw=black,fill=cyan!30, fill opacity=0.2,inner sep=0,fit to=tree]{};}
               [on the floor,roof]
             ]
           ]
         ]
       ]
      ]
    ]
  ]
]
\end{forest}
      }
      \caption{PP recursion}\label{fig:structural-generalization:PPrecursion}
  \end{subfigure}
  \caption{Structural generalization in COGS.}
  \label{fig:structural-generalization}
\end{figure}
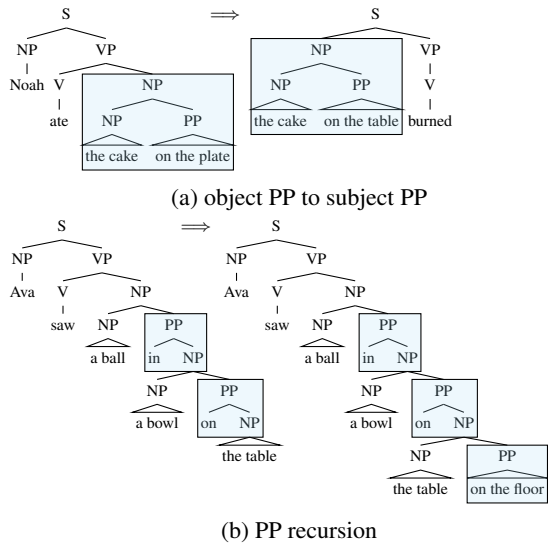

COGS \cite{kim-linzen-2020-cogs} is a synthetic semantic parsing
dataset in which English sentences must be mapped to logic-based
meaning representations (see Fig.~\ref{fig:cogssamples} for some examples). It distinguishes 21 \emph{generalization
  types}, each of which requires generalizing from training instances
to test instances in a particular systematic and
linguistically-informed way. COGS was designed to measure \emph{compositional generalization}, the ability of a semantic parser to assign correct meaning representations to out-of-distribution sentences. Unlike SCAN and CFQ, it includes generalization types with unbounded recursion and separates them cleanly from other generalization types, both of which are crucial for the experiments reported here.

Most generalization types in COGS are \emph{lexical}: they recombine known grammatical structures with words that were not observed in these particular
structures in training.  An example is the generalization type
``subject to object'' (\cref{fig:cogssamples}a), in which a
 noun (``hedgehog'') is only seen as a subject in training,
whereas it is only used as on object at test time. The syntactic structure at test time was already observed in training; only the words change.


By contrast, \emph{structural generalization} involves generalizing to
linguistic structures that were not seen in training (cf.\
\cref{fig:cogssamples}b,c). Examples are the generalization types ``PP
recursion'', where training instances contain prepositional phrases of
depth up to two and generalization instances have PPs of depth 3--12;
and ``object PP to subject PP'', where PPs modify only objects in
training and only subjects at test time. These structural changes are
illustrated in 
\cref{fig:structural-generalization}.

Structural generalization requires learning about recursion and compositionality, and is thus a more thorough test of human-like language use, whereas lexical generalization amounts to smart template filling. In this paper, we investigate how well structural generalization can be solved by different classes of model architectures: \textit{seq2seq models} and \textit{\linf\ models}. We define a model as ``\linf''
if it is explicitly designed to encode linguistic knowledge beyond the fact that sentences are sequences of tokens. This captures a large class of models that can be as ``deep'' as a compositional semantic parser or as ``shallow'' as a POS tagger that requires that each input token gets exactly one POS tag.


\section{Structural generalization is hard for seq2seq}
\label{sec:semantic-parsing}

We begin with some evidence that structural generalization in COGS is hard for seq2seq models, while \linf\ models learn it quite easily. We first collect some results on the original semantic parsing task of COGS, extending it with numbers for BART and T5. We then transform COGS into a corpus for syntactic parsing and POS tagging and investigate the ability of BART and T5 to generalize structurally on these tasks.

\subsection{Experimental setup: COGS}\label{subsec:first:setup}

We follow standard COGS practice and evaluate all models on the generalization set.
We report exact match accuracies, averaged across 5 training runs.

\paragraph{Seq2seq models.} \label{subsec:seq2seq:setup}
We train BART \cite{lewis-etal-2020-bart-acl} and T5 \cite{raffel-etal-2020-t5} as semantic parsers on COGS. Both models are strong representatives of seq2seq models and perform well across many NLP tasks.
To apply these models on COGS, we directly fine-tune the pretrained
\textit{bart-base} and \textit{t5-base} model on it with the corresponding tokenizer; see Appendix \ref{appendix:sec:train} for details.
We also report results for a wide range of published seq2seq models for COGS
\citep{kim-linzen-2020-cogs,conklin-etal-2021-meta,csordas-etal-2021-devil,akyurek-andreas-2021-lexicon, zheng-lapata-2022-disentangled, qiu2021improving}.

\paragraph{\Linf\ models.}  We report evaluation results for LeAR \citep{liu-etal-2021-learning} and the AM parser \citep{weissenhorn22starsem}. Both models learn to predict a tree structure which is decoded into COGS meaning representations using the Principle of Compositionality. Thus both models are \linf.

\begin{table*}
      \centering 
      \scriptsize 
      \setlength{\tabcolsep}{2pt}
      
\begin{tabular}{@{}lll|ccc|c|c@{}}
\toprule
& &  & \multicolumn{3}{c|}{\structg}  & \lexg & \\
& Model Class & Model & Obj to Subj PP & CP recursion & PP recursion & all 18 other types & Overall \\ 
\midrule
\multirow{11}{*}{semantics\qquad} 
& \multirow{9}{*}{seq2seq} & BART  & \asbar{0} & \asbar{0} & \asbar{12} & \asbar{91} & \asbar{79} \\
& & BART+syn  & \asbar{0} & \asbar{5} & \asbar{8} & \asbar{93} & \asbar{80} \\ 
& & T5  & \asbar{0} & \asbar{0} & \asbar{9}  & \asbar{97} & \asbar{83} \\ 
& &  \citealt{kim-linzen-2020-cogs} & \asbar{0} & \asbar{0} & \asbar{0} &  \asbar{73} & \asbar{63} \\
& &  \citealt{akyurek-andreas-2021-lexicon}  & \asbar{0} & \asbar{0} & \asbar{1} & \asbar{96} & \asbar{82} \\
& &  \citealt{zheng-lapata-2022-disentangled}  & \asbar{0} & \asbar{12} & \asbar{39} & \asbar{99} & \asbar{89} \\
& & \citealt{conklin-etal-2021-meta} & \asbar{0} & \asbar{0} & \asbar{0} &\asbar{88} & \asbar{75} \\
& &  \citealt{csordas-etal-2021-devil} & \asbar{0} & \asbar{0} & \asbar{0} & \asbar{95} & \asbar{81} \\
& &  \citealt{qiu2021improving} * & \asbar{100} & \asbar{100} & \asbar{100} & \asbar{100} & \asbar{100} \\
\cmidrule[0.5pt]{2-8} 
& \multirow{2}{*}{\linf\qquad} & \citealt{liu-etal-2021-learning} & \asbar{93} & \asbar{100} & \asbar{99} & \asbar{99} & \asbar{99} \\ 
&  & \citealt{weissenhorn22starsem} & \asbar{78} & \asbar{100} & \asbar{99} & \asbar{100} & \asbar{98} \\ 
\midrule 
\multirow{3}{*}{syntax} & \multirow{2}{*}{seq2seq} & BART  & \asbar{0} & \asbar{9} & \asbar{22} & \asbar{99} & \asbar{87} \\
& &T5  & \asbar{5} & \asbar{7} & \asbar{9} & \asbar{99} & \asbar{86} \\ 
\cmidrule[0.5pt]{2-8} 
& \linf & Neural Berkeley Parser  & \asbar{84} & \asbar{95} & \asbar{98} & \asbar{100} & \asbar{99} \\
\midrule
\multirow{3}{*}{POS tags} & \multirow{2}{*}{seq2seq} & BART & \asbar{0} & \asbar{6} & \asbar{19}  & \asbar{98} & \asbar{85} \\
& & T5 & \asbar{0} & \asbar{4} & \asbar{4} & \asbar{98} & \asbar{85} \\ 
\cmidrule[0.5pt]{2-8} 
& \linf & most frequent POS  & \asbar{92} & \asbar{98} & \asbar{100} & \asbar{92} & \asbar{93} \\
\bottomrule
\end{tabular} 
    \caption{Exact match accuracies on the individual generalization types. 
    Column \lexg\ reports mean accuracy over the 18 lexical generalization types.
    *) After \linf\ data augmentation.
    }\label{tab:selected_gentype_eval}
  \end{table*}


\subsection{Results}
We report the results by generalization type in the ``semantic'' rows in  \Cref{tab:selected_gentype_eval}. We will explain ``BART+syn'' in
\cref{subsec:seq2seq:semantic_gen_with_syntax} and the ``syntactic'' and ``POS''
sections in \cref{subsec:seq2seq:syntactic_gen}.

\paragraph{Structural generalization is hard.} We can observe that all recent models achieve near-perfect accuracy on the 18 lexical generalization types. However, all pure seq2seq models achieve very low accuracy on the structural generalization types, whereas \linf\ models are still very accurate.
One outlier is the seq2seq model of \citet{qiu2021improving}. It employs heavy data augmentation based on (\linf) synchronous grammars encoding the Principle of Compositionality, which provides training instances of higher recursive depth to the seq2seq model. The seq2seq model then still generalizes to the recursive depth which it has seen in training, but not beyond (Peter Shaw, p.c.).

Note that the mean accuracy is dominated by the lexical generalization types; to really measure the ability of a model to generalize to unseen structures, it is important to focus on the structural generalization types. Note further that BART and T5 perform very well among the class of seq2seq models, outperforming many models that are specialized to COGS. We will focus on these two models in the experiments below.

It is important that although the generalization instances on PP and CP recursion are longer than the training instances, the low accuracy of the seq2seq models cannot be explained exclusively in terms of their known weakness to length generalization \citep{hupkes2020compositionality}. For the ``Object to Subject PP'' generalization type, the generalization and training sentences have the same length, but different structures. Thus our results point towards a specific weakness to structural generalization.



\begin{figure}
    \centering
    \includegraphics[scale=.3,trim={5mm 5mm 40mm 0},clip]{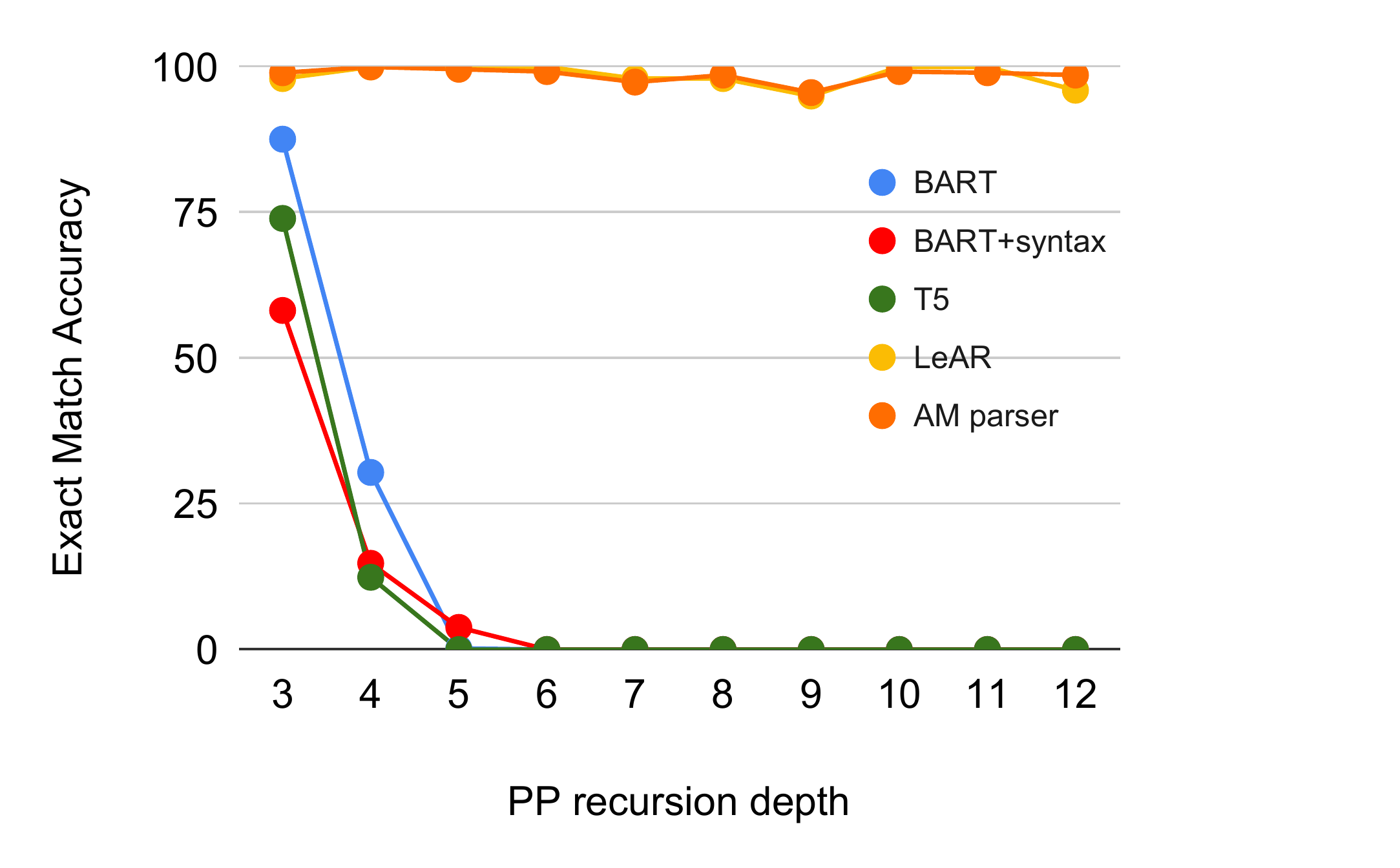}
    \caption{Influence of PP recursion depth on overall PP depth
      generalization accuracy. 
    }\label{fig:depths_plot}
  \end{figure}

\paragraph{Depth generalization.}
The accuracy of the seq2seq models depends on the difference in complexity of the test instance and the training data. For instance, all training instances for the ``PP recursion'' type have recursion depth two or less; \cref{fig:depths_plot} shows how the accuracy depends on the recursion depth of the test instance. As we see,
the accuracy of BART (even when informed by syntax, cf.\
\cref{subsec:seq2seq:semantic_gen_with_syntax}) degrades
quickly with recursion depth. By contrast, LeAR and the AM parser
maintain high accuracy across all recursion depths.


\subsection{\syncogs\ and \poscogs} \label{sec:syntax}
\label{subsec:seq2seq:syntactic_gen}
While these results on semantic parsing are suggestive, they could be explained away in many ways. For instance, the weakness of seq2seq models with respect to structural generalization might be specific to semantics, or the semantic representations chosen in COGS might be idiosyncractic and unfair to seq2seq models.

We therefore investigate structural generalization on a syntax variant of COGS. We convert each training and generalization instance of the COGS corpus into a pair of the sentence with its syntax tree (\syncogs) and a pair of the sentence with its POS tag sequence (\poscogs). This is possible because COGS is generated from an unambiguous context-free grammar; we reconstruct the unique syntax trees that underly each instance in COGS.

We replace the very
fine-grained non-terminals (e.g.\  \lform{NP\_animate\_dobj\_noPP}) of
the original COGS grammar with more general ones (e.g.\ \lform{NP}) and remove
duplicate rules (e.g.\ \lform{NP}$\to$\lform{NP}) resulting from this. 
We extract the POS tag sequences from the preterminal nodes of the syntax trees.

We train BART and T5 to predict linearized constituency trees 
and the POS tag sequences from the
input sentences. As a \linf\ model, we use the 
Neural Berkeley
Parser \citep{kitaev-klein-2018-constituency}, which consists of a self-attention
encoder and a chart decoder and therefore has the notion of a tree and its recursive structure built into the parsing model. 
On the POS tagging task, our ``\linf'' model is constrained to predict exactly one POS tag for each input token. Specifically, we determine the most frequent POS tag in the training data for each word type and assign it to all occurrences of the word during inference.


\paragraph{Results.}
The results are shown in the  ``syntactic'' and  ``POS'' rows of
\cref{tab:selected_gentype_eval}. We find the same pattern as in the
semantic parsing case: the seq2seq models do well on
\lexg, but struggle with \structg. The \linf\ models 
handle all generalization types well.
Thus, the
difficulties that seq2seq models have on structural generalization on
COGS are not limited to semantics: rather, they seem to be a general
limitation in the ability of seq2seq models to learn linguistic
structure from structurally simple examples and use it
productively. 



\begin{figure}
    \centering
    \scriptsize
    \begin{tabular}{@{}llm{.7\linewidth}@{}}
        \toprule
         &  {Input} & {The baby on a tray in the house screamed.}  \\
         \midrule
         \multirow{2}{*}{ \rotatebox[origin=c]{90}{Semantics\quad}} &  {Gold} &  {\lform{*baby($x_1$); *house($x_7$); baby.nmod.on($x_1$,$x_4$) $\land$ tray($x_4$) $\land$ tray.nmod.in($x_4$,$x_7$) $\land$ scream.agent($x_8$,$x_1$) }} \\
         \cmidrule(l){2-3}
         &  {T5} &  {\lform{*baby($x_1$); *house(\textcolor{red}{$x_{10}$}); \textcolor{red}{scream.agent($x_2$,$x_1$)} $\land$ \textcolor{red}{scream.theme($x_2$,$x_4$)} $\land$ tray($x_4$) $\land$ tray.nmod.in($x_4$,$x_7$)}} \\
         \midrule
         \multirow{2}{*}{\rotatebox[origin=c]{90}{Syntax\quad\quad}} &  {Gold} & {\lform{( S ( NP ( Det The ) ( N baby ) ( PP ( P on ) ( NP ( Det a ) ( N tray ) ( PP ( P in ) ( NP ( Det the ) ( N house ) ) ) ) ) ) ( VP ( V screamed ) ) ) }} \\
         \cmidrule(l){2-3}
         &  {T5} & {\lform{( S ( NP ( Det The ) ( N baby ) ( \textcolor{red}{VP} ( \textcolor{red}{V} on ) ( NP ( Det a ) ( N tray ) ( PP ( P in ) ( NP ( Det the ) ( N house ) ) ) ) ) ) \textcolor{red}{)}}} \\
         \midrule
         \multirow{2}{*}{\rotatebox[origin=c]{90}{POS\;\;}} &  {Gold} & {\lform{Det N P Det N P Det N V}} \\
         \cmidrule(l){2-3}
         &   {T5} & {\lform{Det N \textcolor{red}{V} Det N P Det N \textcolor{red}{P Det N}}} \\
         \bottomrule
    \end{tabular} 
    \caption{Example for \textit{obj\_to\_subj\_pp} type. We list the annotation of semantic parse, syntax tree and POS tags with corresponding T5 predictions. }
    \label{fig:error_example}
\end{figure}

We also present an example for \textit{obj\_pp\_to\_subj\_pp} type across different tasks in Figure \ref{fig:error_example}. For a sentence \textit{The baby on a tray in the house screamed}, T5 consistently predicted wrong symbol sequences. For example, in semantic parsing, T5 tends to predict \textit{tray} as the theme of \textit{scream} with a PP structure. This might be due to a preference of T5 to reuse the pattern for object-PP sentences in the train set even if the intransitive verb does not license it. T5 also displays an unawareness
of word order that is reminiscent of the difficulties that seq2seq models otherwise face in relating syntax to word order \citep{DBLP:journals/tacl/McCoyFL20}. For recursion generalization types, we find that the main error is that the decoder cannot generate long or deep enough sequences. 


\section{Encoder or decoder?}
\label{sec:probing}

We now turn to the second question: \emph{Why} do seq2seq models struggle on structural generalization? We start by investigating at which point the model loses the structural information -- does the encoder not represent it, or can the decoder not make use of it? This also addresses an apparent tension between our findings and previous work demonstrating that pretrained models contain rich linguistic information \cite{hewitt-manning-2019-structural,DBLP:conf/iclr/TenneyXCWPMKDBD19}, which should be sufficient to at least solve \syncogs.


\subsection{Probing for structural information}
\label{sec:probing-howto}
We use the well-established probe task methodology \cite{peters-etal-2018-dissecting,tenney-etal-2019-bert} to analyze what information is present in the outputs of the BART encoder. We define both a syntactic and a semantic probing task:

\paragraph{Constituent labeling.} The goal of this task is to predict correct labels for all constituency spans in a sentence. We treat spans that are not constituents as if they were annotated with the \textit{None} label. The gold annotations are derived from \syncogs.


\paragraph{Semantic role labeling.} To measure the presence of structural semantic information, we define a probe task that predicts role labels for all predicate-argument relations in a sentence. For example, in the sentence \textit{Emma slept}, the goal is to recognize that \textit{slept} is a predicate with \textit{Emma} being its \textit{agent}. This task captures most of the information in the original COGS meaning representations as relations between tokens in the sentence. We extract data for this task (given two tokens, predict if the second is an argument of the first and with what role label) from the COGS meaning representation. We refer to  Appendix \ref{appendix:sec:srl} for details.

We train probe classifiers in a similar way as \cite{tenney-etal-2019-bert}. For each task, we train a multi-layer perceptron to predict the target label from the outputs of the frozen pretrained encoder. For constituent labeling, 
the MLP reads a span representation obtained by subtracting the encodings of the tokens at the span boundary from each other \cite{stern-etal-2017-minimal}.  For semantic role labeling, the input of the MLP is the concatenation of the encodings for the predicate and argument token.

We evaluate the probes in two ways. First, we train the probes on the original training split of COGS  (``orig'').  However, this conflates the presence of structural information in the encodings with the ability of the probing MLP itself to perform structural generalization. We therefore also evaluate on a second split (``probe'') in which we add 60\% of the generalization set (randomly selected) to the training set and 10\% to the development set and keep the rest as the probe test set. This makes the probe test set in-distribution with respect to the probe training set. The encoder remains frozen and can therefore not adapt to the modified training set; we still obtain meaningful results about whether the pretrained encodings contain the information that is needed to learn to predict structure in COGS.


\subsection{Results}
We report the sentence-level accuracy in Table \ref{tab:probe_eval}. 
For better comparison, all accuracies are measured on the test set from the ``probe'' split.
We find that the probes learn to solve both tasks accurately on the ``probe'' split, indicating that the pretrained encodings of BART contain all the information that is needed to make structural predictions. By contrast, when we replace the BART encodings with random vectors of the same size (``Random'' rows), the probe fails to learn. The probes also perform badly on the ``orig'' split, suggesting that the probe ``decoder'' does not generalize structurally either.

These findings suggest that the BART encoder captures all the necessary information about the input sentence, but the BART decoder cannot use it to learn to generalize structurally.



\begin{table}
      \centering 
      \scriptsize 
      \setlength{\tabcolsep}{2pt}
      
\begin{tabular}{@{}lll|ccc|c@{}}
\toprule
& &  & \multicolumn{3}{c|}{\structg} & \lexg \\
& Encoder & Data & Obj to Subj PP & CP recursion & PP recursion  & all 18 other types  \\ 
\midrule
\parbox[t]{1mm}{\multirow{2}{*}{\rotatebox[origin=c]{90}{sem}}} 

& BART & probe & \asbar{82} & \asbar{91} & \asbar{92}  & \asbar{100}  \\
& Random & probe & \asbar{25} & \asbar{0} & \asbar{65}  & \asbar{90}  \\
& BART & orig & \asbar{0} & \asbar{5} & \asbar{27} & \asbar{94} \\
\midrule
\parbox[t]{1mm}{\multirow{2}{*}{\rotatebox[origin=c]{90}{syn}}} & BART & probe & \asbar{85} & \asbar{80} & \asbar{83} & \asbar{100}  \\
& Random & probe & \asbar{1} & \asbar{0} & \asbar{0}  & \asbar{16}  \\
& BART & orig & \asbar{0} & \asbar{0} & \asbar{7} & \asbar{92} \\
\bottomrule
\end{tabular} 
    \caption{Exact match accuracy for probing on the individual generalization types. 
    }\label{tab:probe_eval}
  \end{table}


\subsection{Enriching seq2seq with structure}\label{subsec:seq2seq:semantic_gen_with_syntax}

Can we make things easier for the decoder by making the structural information explicit in the input? To investigate this, we inject the gold syntax tree into the BART encoder to see if this improves structural generalization in semantic parsing.


We retrain BART on COGS, but
instead of feeding it the raw sentence, we provide as input the
linearized gold constituency tree (``\lform{(NP ({Det} a) ({N} rose))}''),
both for training and inference.
 This method is similar to 
\citet{li-etal-2017-modeling} and \citet{currey-heafield-2019-incorporating},
but we allow attention over special tokens such as ``\lform{(}'' during decoding.

We report the results as ``BART+syn'' in
\cref{tab:selected_gentype_eval} and \cref{fig:depths_plot}; the overall accuracy increases by
1.5\% over BART. This
is mostly because providing the syntax tree allows BART to generalize
correctly on \lexg. 
However, \structg remains out
of reach for BART+syn, confirming the deep difficulty of
structural generalization for seq2seq models.

We also explored other ways to inform BART with syntax, through
multi-task learning \cite{sennrich-etal-2016-controlling,
  currey-heafield-2019-incorporating} and syntax-based masking in the
self-attention encoder \cite{kim-etal-2021-improving}. Neither method
substantially improved the accuracy of BART on the COGS generalization
set (+1.0\% and -6.4\% overall accuracy, respectively). 
We conclude that the weakness of the BART decoder towards structural generalization persists even when the input makes the structure explicit.

\section{Text-to-text structural generalization} \label{sec:overview}

We will now turn our attention to a novel text-to-text variant of COGS. The difficulty of structural generalization for seq2seq models has been primarily studied on tasks where sentences must be mapped into symbolic representations of some kind, such as the semantic and syntactic representations in Section~\ref{sec:semantic-parsing}. But although pretrained seq2seq models like BART and T5 achieve excellent accuracy on broad-coverage semantic parsing tasks, one might argue that they were originally designed for tasks where the output sequence is natural language as well, and thus should be evaluated on such tasks.

We therefore propose a new dataset, \cogsqa, which presents structural generalization examples based on COGS sentences in a question-answering format. Given a context sentence and a question sentence as input, the goal is to output the correct answer, which should be a consecutive span of tokens in the context sentence. The dataset consists of two sections: \emph{\cogsqabase} directly asks questions about COGS sentences (\cref{subsec:arg_rel}), whereas \emph{\cogsqacc} combines COGS sentences in novel coordinating structures (\cref{subsec:cc_cp}). Following the original COGS design, each section consists of four subsets: training set, development set, in-distribution test set, and out-of-distribution generalization set.

\begin{figure}
    \centering
    \includegraphics[scale=0.6]{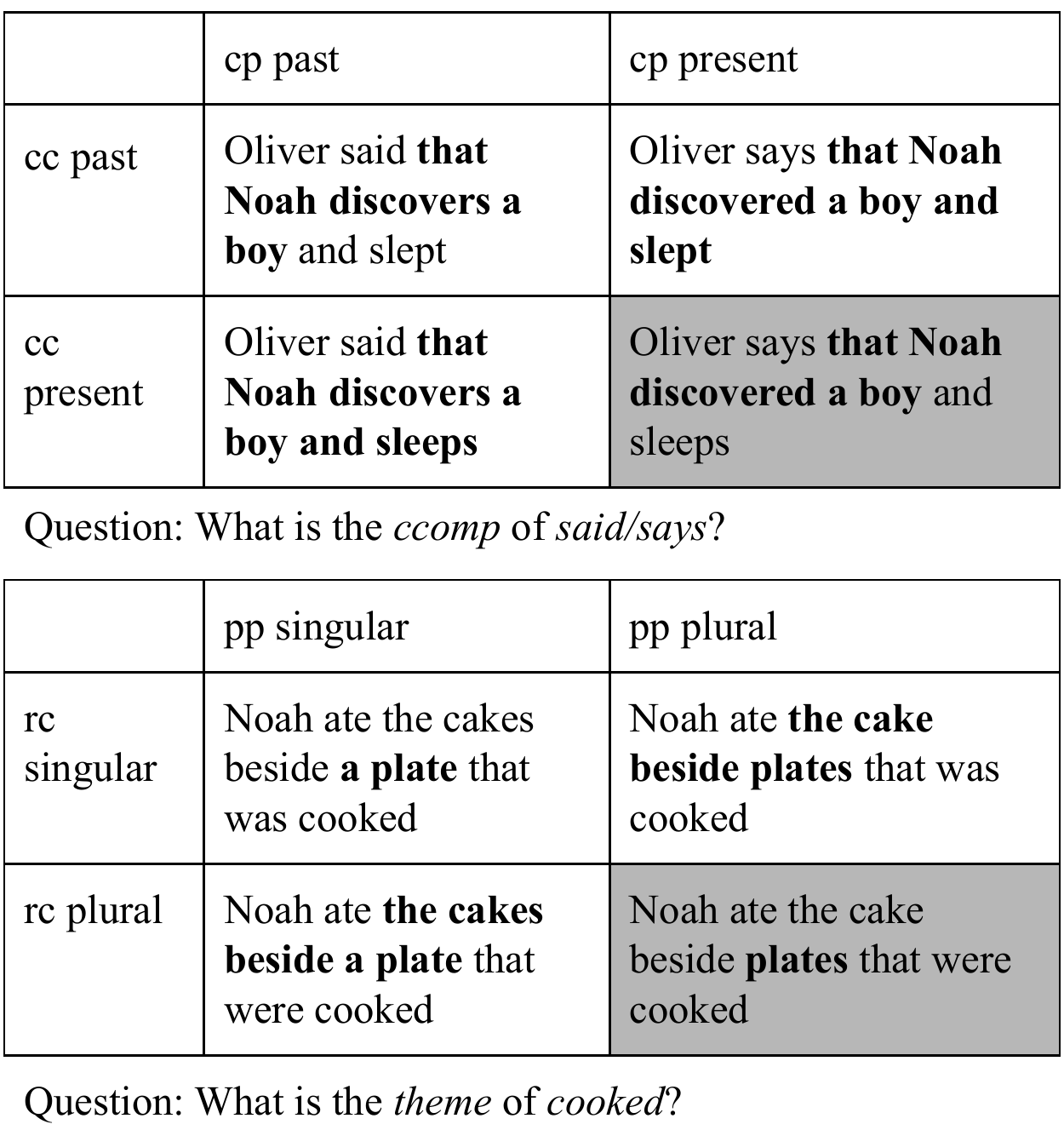}
    \caption{Construction of \cogsqacc: top is \textit{cc\_cp}, bottom is \textit{rc\_pp}. The answer to the example question is highlighted in bold.
 }
    \label{fig:cc_cp_example}
\end{figure}



\begin{figure*}[htbp!]
    \small
    \centering
    \setlength{\tabcolsep}{2pt}
    \begin{tabularx}{\linewidth}{lXX}
        \toprule
        \textbf{Gen type} & \textbf{Training}  &  \textbf{Generalization} \\
       \midrule
        {Obj to Subj PP}  & {Noah ate  {\textbf{the cake on the plate}}. What did Noah eat?} & { {\textbf{The cake on the plate}} burned. What was burned?}\\ 
        \midrule 
        {PP recursion}  & {Ava saw {\textbf{a ball in a bowl on the table}}. What did Ava see?} & {Ava saw  {\textbf{a ball in a bowl on the table on the floor}}. What did Ava see?} \\ 
        \midrule
        {CP recursion}  & {Ava said  {\textbf{that Emma liked that a dog ran}}. What did Ava say?} & {Ava said  {\textbf{that Emma liked that Noah noticed that a dog ran}}. What did Ava say?} \\ 
        \bottomrule

    \end{tabularx}
    \caption{Examples for the \cogsqabase\  dataset with regard to each structural generalization type. In each example, the first sentence is the context sentence, the second sentence is the question sentence and the bold token span is the corresponding answer.}
    \label{fig:qa_gen_examples}
\end{figure*}

\subsection{\cogsqabase} \label{subsec:arg_rel}
The \cogsqabase\ dataset uses the sentences of COGS as context sentences, and then asks one or more questions about each sentence that can be answered by a contiguous substring (see Fig.\ \ref{fig:qa_gen_examples}). For example, given \textit{Noah ate the cake on the plate} as context, we ask \textit{What did Noah eat?} and \textit{Who ate the cake on the plate?}, and the answer should be \textit{the cake on the plate} and \textit{Noah} respectively.

To generate question-answer pairs, we identify the semantic roles and arguments for each predicate in all sentences of COGS, as in the SRL probing task (\cref{sec:probing-howto}). We generate question-answer pairs out of these based on handwritten templates (i.e.\ at least one per COGS instance) and split them into train/test/generalization sets as in the original COGS. We refer to Appendix \ref{appendix:sec:qa} for more details.

The original COGS training set contains ``primitive'' instances in which the sentence consists of a single word, and the meaning representation is the word itself (e.g.\ Paula $\Rightarrow$ Paula). We include these instances in \cogsqabase\ by using a special token \textit{<prim>} as the question sentence and the primitive word as context and answer (i.e., Paula \textit{<prim>} $\Rightarrow$ Paula).

\begin{table*}[htbp!]
      \centering 
      \scriptsize 
      \setlength{\tabcolsep}{2pt}
      
\begin{tabular}{@{}ll|ccc|c|c|c|c@{}}
\toprule
 & & \multicolumn{5}{c|}{\cogsqabase} & \multicolumn{2}{c}{\cogsqacc} \\
&   & \multicolumn{3}{c|}{\structg}  & \lexg & & & \\
Model Class & Model & Obj to Subj PP & CP recursion & PP recursion & all 18 other types & Overall & \textit{cc\_cp} & \textit{rc\_pp} \\ 
\midrule
\multirow{2}{*}{seq2seq} & BART  & \asbar{99} & \asbar{59} & \asbar{69} & \asbar{95} & \asbar{86} & \asbar{37} & \asbar{14} \\
& T5  & \asbar{100} & \asbar{95} & \asbar{97} & \asbar{100} & \asbar{99} & \asbar{16} & \asbar{22} \\ \midrule
\multirow{2}{*}{\linf\qquad} & BART-QA  & \asbar{100} & \asbar{98} & \asbar{100}  & \asbar{100} & \asbar{99} & \asbar{6} & \asbar{0}\\
& BART-QA+struct\qquad\  & {-} & {-} & {-}  & {–} & {-} & \asbar{100} & \asbar{100}\\
\bottomrule
\end{tabular} 
    \caption{Exact match accuracy on the individual generalization types on the sections of \cogsqa.
    }\label{tab:qa_gentype_eval}
  \end{table*}

\subsection{\cogsqacc} \label{subsec:cc_cp}
We add \cogsqacc\ as a second, harder text-to-text task based on COGS. This task exploits the interplay of the syntactic structure of a sentence with constraints on tense and number agreement. For instance, in sentences of the form ``N1 V1 that N2 V2 and V3'' (where N1, N2 are noun phrases and V1, V2, V3 are verbs),
V3 belongs to the same clause as V1 or V2 depending on which one it agrees with. Thus, the agreement between verbs disambiguates a structural ambiguity of the sentence. Some concrete examples are shown in Fig.~\ref{fig:cc_cp_example}. The idea that agreement interacts with syntax is reminiscent of \citet{linzen2016assessing}, but here we predict the syntactic structure rather than the agreement feature.

\cogsqacc\ consists of two parts. The subcorpus \textit{cc\_cp} consists of sentences as above, where tense agreement disambiguates the structural ambiguity between CP embedding and coordination. The subcorpus \textit{rc\_pp} contains sentences where number agreement disambiguates the attachment of a relative clause. In both cases, we construct context sentences using a context-free grammar adapted from the one that generates COGS. We generate questions of the form ``What is the ccomp of said?'' along with their answers from the context sentences using a small number of hand-written heuristics.  Answering these questions correctly amounts to disambiguating the structure of the sentence.

We create training (4k instances), development (1k), and in-domain test sets  (1k) for \cogsqacc\ out of three of the four combinations of the agreement features of the two verbs (white cells in Fig.~\ref{fig:cc_cp_example}). We create a generalization set (2k instances) from the fourth, unseen combination of agreement features (gray cells).

\section{Experiments on \cogsqa} \label{sec:qa_experiments}

\subsection{Models} \label{subsec:seq2seq:setup}

We conduct a series of experiments in which a model receives the concatenation of context sentence and question as input and must predict the answer. 
We fine-tune BART and T5 on \cogsqa\ and compare against two \linf\ models. Details of the training setup are discussed in Appendix \ref{appendix:sec:train}.

First, we compare against an extractive model we call BART-QA. Given a context sentence and question, BART-QA predicts the start and end position of the answer within the context sentence. The start and end positions are each predicted by an MLP trained from scratch which takes the outputs of the pretrained BART encoder as input.


Second, we use a more informed model called BART-QA+struct specifically for \cogsqacc. {BART-QA+struct} shares the same encoder as {BART-QA}, but its decoder is constrained to select a span which exists in the gold syntax tree of the sentence. This model accesses information that is usually not available at test time, and we offer it only as a point of comparison.



\subsection{Results}

The exact match accuracies on the generalization sets are shown in Table \ref{tab:qa_gentype_eval}. Similar to the earlier experiments, all models perform well on \lexg; we mainly discuss results on \structg below. 

\paragraph{\cogsqabase.} All models solve ``Object to Subject PP'' perfectly, with T5 and BART-QA also achieving perfect accuracy on the PP and CP recursion. While these positive results on structural generalization seem to go against the grain of our earlier discussion, it is important to note that \cogsqabase\ is an extractive task which only requires selecting a substring of the input; and further, that this substring is in a very specific position of the string, making the task amenable to learning simple heuristics (e.g.\ subject is everything to the left of the verb). Thus, these results indicate that structural generalization is hard only if the decoder's task is sufficiently complex. Note that unlike BART, T5 sees question answering tasks during training, which may help explain the difference in accuracy.

\paragraph{\cogsqacc.} However, BART and T5 all achieve low accuracy on \cogsqacc, suggesting that even text-to-text tasks involving structural generalization can be difficult; string-level heuristics are not successful on this task. In this case, the task is still hard for the \linf\ model BART-QA. It can be solved by BART-QA+struct, but note that this model has access to gold syntax information which makes the task much easier.
Note that since the training and generalization sentences in \cogsqacc\ are of similar length, the difficulty comes exclusively from structural rather than length generalization.

\section{Conclusion}
We have presented evidence that structural generalization is hard for seq2seq models, both on semantic and syntactic parsing (COGS and \syncogs) and on some text-to-text tasks (\cogsqacc). In many of these cases, \linf\ models generalize successfully where seq2seq models struggle. Unlike earlier work, we have shown that this effect persists when the seq2seq models can be pretrained.

We have then presented a number of experiments to help pinpoint the cause of this limitation. We found that the BART encoder still provides structural information, but the decoder does not use it to generalize -- both in the parsing tasks and in the probing tasks on the original splits, and not even when the input is enriched with syntactic information. We further found that when the decoder's task is simple enough, as in \cogsqabase, seq2seq models learn to generalize structurally as well as \linf\ models. In improving the ability of seq2seq models to generalize structurally, it seems promising to focus on the decoder, especially by including \linf\ elements.

\section{Limitations} 
Our experiments are limited to a synthetic corpus (COGS) and its derivatives. While it seems plausible to us to justify negative results like ours with a synthetic corpus, it must be recognized that the distribution of language in COGS is not the same as in English as a whole, which might undermine the ability of both seq2seq and \linf\ models to learn to generalize.

Furthermore, claims about a whole class of models (seq2seq) can only be supported, never completely proved, through empirical experiments on a finite set of representatives. Nonetheless, we think that this paper has considered a sufficiently wide range of models and tasks to make careful statements about seq2seq models as a class.

\section*{Acknowledgements}
We are indebted to Lucia Donatelli and Pia Weißenhorn for fruitful discussions, and to Najoung Kim for providing the code for generating COGS.
This work was supported by the Deutsche Forschungsgemeinschaft (DFG) through the project KO 2916/2-2.

\bibliography{anthology,custom}

\begin{thebibliography}{35}
\expandafter\ifx\csname natexlab\endcsname\relax\def\natexlab#1{#1}\fi

\bibitem[{Akyürek and Andreas(2021)}]{akyurek-andreas-2021-lexicon}
Ekin Akyürek and Jacob Andreas. 2021.
\newblock \href {https://aclanthology.org/2021.acl-long.382} {Lexicon learning
  for few shot sequence modeling}.
\newblock In \emph{Proceedings of the 59th Annual Meeting of the Association
  for Computational Linguistics and the 11th International Joint Conference on
  Natural Language Processing (Volume 1: Long Papers)}, pages 4934--4946,
  Online. Association for Computational Linguistics.

\bibitem[{Bevilacqua et~al.(2021)Bevilacqua, Blloshmi, and
  Navigli}]{bevilacqua-etal-2021-one}
Michele Bevilacqua, Rexhina Blloshmi, and Roberto Navigli. 2021.
\newblock \href {https://ojs.aaai.org/index.php/AAAI/article/view/17489} {One
  {SPRING} to rule them both: {S}ymmetric {AMR} semantic parsing and generation
  without a complex pipeline}.
\newblock In \emph{Proceedings of the AAAI Conference on Artificial
  Intelligence (AAAI-21)}, volume~35, pages 12564--12573. AAAI Press.

\bibitem[{Bogin et~al.(2022)Bogin, Gupta, and
  Berant}]{bogin-2022-local-structure}
Ben Bogin, Shivanshu Gupta, and Jonathan Berant. 2022.
\newblock \href {https://doi.org/10.48550/ARXIV.2201.05899} {Unobserved local
  structures make compositional generalization hard}.
\newblock \emph{arXiv preprint arXiv:2201.05899}.

\bibitem[{Chomsky(1957)}]{chomsky-1957-syntactic}
Noam Chomsky. 1957.
\newblock \href {https://doi.org/10.1515/9783112316009} {\emph{Syntactic
  Structures}}.
\newblock De Gruyter Mouton.

\bibitem[{Conklin et~al.(2021)Conklin, Wang, Smith, and
  Titov}]{conklin-etal-2021-meta}
Henry Conklin, Bailin Wang, Kenny Smith, and Ivan Titov. 2021.
\newblock \href {https://aclanthology.org/2021.acl-long.258} {Meta-learning to
  compositionally generalize}.
\newblock In \emph{Proceedings of the 59th Annual Meeting of the Association
  for Computational Linguistics and the 11th International Joint Conference on
  Natural Language Processing (Volume 1: Long Papers)}, pages 3322--3335,
  Online. Association for Computational Linguistics.

\bibitem[{Csord{\'a}s et~al.(2021)Csord{\'a}s, Irie, and
  Schmidhuber}]{csordas-etal-2021-devil}
R{\'o}bert Csord{\'a}s, Kazuki Irie, and J{\"u}rgen Schmidhuber. 2021.
\newblock \href {https://aclanthology.org/2021.emnlp-main.49} {The devil is in
  the detail: Simple tricks improve systematic generalization of transformers}.
\newblock In \emph{Proceedings of the 2021 Conference on Empirical Methods in
  Natural Language Processing}, pages 619--634, Online and Punta Cana,
  Dominican Republic. Association for Computational Linguistics.

\bibitem[{Currey and Heafield(2019)}]{currey-heafield-2019-incorporating}
Anna Currey and Kenneth Heafield. 2019.
\newblock \href {https://doi.org/10.18653/v1/W19-5203} {Incorporating source
  syntax into transformer-based neural machine translation}.
\newblock In \emph{Proceedings of the Fourth Conference on Machine Translation
  (Volume 1: Research Papers)}, pages 24--33, Florence, Italy. Association for
  Computational Linguistics.

\bibitem[{Fodor and Lepore(2002)}]{fodor-lepore-2002-compositionality}
Jerry~A. Fodor and Ernest Lepore. 2002.
\newblock \emph{The Compositionality Papers}.
\newblock Oxford University Press.

\bibitem[{Fodor and Pylyshyn(1988)}]{fodor-pylyshyn-1988-connectionism}
Jerry~A. Fodor and Zenon~W. Pylyshyn. 1988.
\newblock \href {https://doi.org/10.1016/0010-0277(88)90031-5} {Connectionism
  and cognitive architecture: A critical analysis}.
\newblock \emph{Cognition}, 28(1):3--71.

\bibitem[{Furrer et~al.(2020)Furrer, van Zee, Scales, and
  Schärli}]{furrer-etal-2020-compositional}
Daniel Furrer, Marc van Zee, Nathan Scales, and Nathanael Schärli. 2020.
\newblock \href {http://arxiv.org/abs/2007.08970} {Compositional generalization
  in semantic parsing: Pre-training vs. specialized architectures}.
\newblock \emph{Computing Research Repository (CoRR)}, arXiv:2007.08970.

\bibitem[{Hahn(2020)}]{hahn-2020-theoretical}
Michael Hahn. 2020.
\newblock \href {https://doi.org/10.1162/tacl_a_00306} {Theoretical limitations
  of self-attention in neural sequence models}.
\newblock \emph{Transactions of the Association for Computational Linguistics},
  8:156--171.

\bibitem[{Hewitt and Manning(2019)}]{hewitt-manning-2019-structural}
John Hewitt and Christopher~D. Manning. 2019.
\newblock \href {https://doi.org/10.18653/v1/N19-1419} {{A} structural probe
  for finding syntax in word representations}.
\newblock In \emph{Proceedings of the 2019 Conference of the North {A}merican
  Chapter of the Association for Computational Linguistics: Human Language
  Technologies, Volume 1 (Long and Short Papers)}, pages 4129--4138,
  Minneapolis, Minnesota. Association for Computational Linguistics.

\bibitem[{Hupkes et~al.(2020)Hupkes, Dankers, Mul, and
  Bruni}]{hupkes2020compositionality}
Dieuwke Hupkes, Verna Dankers, Mathijs Mul, and Elia Bruni. 2020.
\newblock \href {https://www.jair.org/index.php/jair/article/view/11674}
  {Compositionality decomposed: how do neural networks generalise?}
\newblock \emph{Journal of Artificial Intelligence Research}, 67:757--795.

\bibitem[{Keysers et~al.(2020)Keysers, Sch{\"a}rli, Scales, Buisman, Furrer,
  Kashubin, Momchev, Sinopalnikov, Stafiniak, Tihon, Tsarkov, Wang, van Zee,
  and Bousquet}]{keysers-etal-2020-measuring}
Daniel Keysers, Nathanael Sch{\"a}rli, Nathan Scales, Hylke Buisman, Daniel
  Furrer, Sergii Kashubin, Nikola Momchev, Danila Sinopalnikov, Lukasz
  Stafiniak, Tibor Tihon, Dmitry Tsarkov, Xiao Wang, Marc van Zee, and Olivier
  Bousquet. 2020.
\newblock \href {https://openreview.net/pdf?id=SygcCnNKwr} {Measuring
  compositional generalization: A comprehensive method on realistic data}.
\newblock In \emph{International Conference on Learning Representations
  (ICLR)}.

\bibitem[{Kim et~al.(2021)Kim, Ravikumar, Ainslie, and
  Ontañón}]{kim-etal-2021-improving}
Juyong Kim, Pradeep Ravikumar, Joshua Ainslie, and Santiago Ontañón. 2021.
\newblock \href {https://doi.org/10.18653/v1/2021.acl-short.81} {Improving
  compositional generalization in classification tasks via structure
  annotations}.
\newblock In \emph{Proceedings of the 59th Annual Meeting of the Association
  for Computational Linguistics and the 11th International Joint Conference on
  Natural Language Processing (Volume 2: Short Papers)}, pages 637--645,
  Online. Association for Computational Linguistics.

\bibitem[{Kim and Linzen(2020)}]{kim-linzen-2020-cogs}
Najoung Kim and Tal Linzen. 2020.
\newblock \href {https://doi.org/10.18653/v1/2020.emnlp-main.731} {{COGS}: A
  compositional generalization challenge based on semantic interpretation}.
\newblock In \emph{Proceedings of the 2020 Conference on Empirical Methods in
  Natural Language Processing}, pages 9087--9105, Online. Association for
  Computational Linguistics.

\bibitem[{Kingma and Ba(2015)}]{DBLP:journals/corr/KingmaB14}
Diederik~P. Kingma and Jimmy Ba. 2015.
\newblock \href {http://arxiv.org/abs/1412.6980} {Adam: {A} method for
  stochastic optimization}.
\newblock In \emph{3rd International Conference on Learning Representations,
  {ICLR} 2015, San Diego, CA, USA, May 7-9, 2015, Conference Track
  Proceedings}.

\bibitem[{Kitaev and Klein(2018)}]{kitaev-klein-2018-constituency}
Nikita Kitaev and Dan Klein. 2018.
\newblock \href {https://doi.org/10.18653/v1/P18-1249} {Constituency parsing
  with a self-attentive encoder}.
\newblock In \emph{Proceedings of the 56th Annual Meeting of the Association
  for Computational Linguistics (Volume 1: Long Papers)}, pages 2676--2686,
  Melbourne, Australia. Association for Computational Linguistics.

\bibitem[{Lake and Baroni(2018)}]{lake-baroni-2018-generalization}
Brenden Lake and Marco Baroni. 2018.
\newblock \href {http://proceedings.mlr.press/v80/lake18a.html} {Generalization
  without systematicity: On the compositional skills of sequence-to-sequence
  recurrent networks}.
\newblock In \emph{Proceedings of the 35th International Conference on Machine
  Learning}, volume~80 of \emph{Proceedings of Machine Learning Research},
  pages 2873--2882, Stockholmsmässan, Stockholm Sweden. PMLR.

\bibitem[{Lewis et~al.(2020)Lewis, Liu, Goyal, Ghazvininejad, Mohamed, Levy,
  Stoyanov, and Zettlemoyer}]{lewis-etal-2020-bart-acl}
Mike Lewis, Yinhan Liu, Naman Goyal, Marjan Ghazvininejad, Abdelrahman Mohamed,
  Omer Levy, Veselin Stoyanov, and Luke Zettlemoyer. 2020.
\newblock \href {https://doi.org/10.18653/v1/2020.acl-main.703} {{BART}:
  Denoising sequence-to-sequence pre-training for natural language generation,
  translation, and comprehension}.
\newblock In \emph{Proceedings of the 58th Annual Meeting of the Association
  for Computational Linguistics}, pages 7871--7880, Online. Association for
  Computational Linguistics.

\bibitem[{Li et~al.(2017)Li, Xiong, Tu, Zhu, Zhang, and
  Zhou}]{li-etal-2017-modeling}
Junhui Li, Deyi Xiong, Zhaopeng Tu, Muhua Zhu, Min Zhang, and Guodong Zhou.
  2017.
\newblock \href {https://doi.org/10.18653/v1/P17-1064} {Modeling source syntax
  for neural machine translation}.
\newblock In \emph{Proceedings of the 55th Annual Meeting of the Association
  for Computational Linguistics (Volume 1: Long Papers)}, pages 688--697,
  Vancouver, Canada. Association for Computational Linguistics.

\bibitem[{Linzen et~al.(2016)Linzen, Dupoux, and
  Goldberg}]{linzen2016assessing}
Tal Linzen, Emmanuel Dupoux, and Yoav Goldberg. 2016.
\newblock \href {https://doi.org/10.1162/tacl_a_00115} {Assessing the ability
  of {LSTM}s to learn syntax-sensitive dependencies}.
\newblock \emph{Transactions of the Association for Computational Linguistics},
  4:521--535.

\bibitem[{Liu et~al.(2021)Liu, An, Lin, Liu, Chen, Lou, Wen, Zheng, and
  Zhang}]{liu-etal-2021-learning}
Chenyao Liu, Shengnan An, Zeqi Lin, Qian Liu, Bei Chen, Jian-Guang Lou, Lijie
  Wen, Nanning Zheng, and Dongmei Zhang. 2021.
\newblock \href {https://aclanthology.org/2021.findings-acl.97} {Learning
  algebraic recombination for compositional generalization}.
\newblock In \emph{Findings of the Association for Computational Linguistics:
  ACL-IJCNLP 2021}, pages 1129--1144, Online. Association for Computational
  Linguistics.

\bibitem[{McCoy et~al.(2020)McCoy, Frank, and
  Linzen}]{DBLP:journals/tacl/McCoyFL20}
R.~Thomas McCoy, Robert Frank, and Tal Linzen. 2020.
\newblock \href {https://transacl.org/ojs/index.php/tacl/article/view/1892}
  {Does syntax need to grow on trees? sources of hierarchical inductive bias in
  sequence-to-sequence networks}.
\newblock \emph{Trans. Assoc. Comput. Linguistics}, 8:125--140.

\bibitem[{Ontanon et~al.(2022)Ontanon, Ainslie, Fisher, and
  Cvicek}]{ontanon-etal-2022-making}
Santiago Ontanon, Joshua Ainslie, Zachary Fisher, and Vaclav Cvicek. 2022.
\newblock \href {https://doi.org/10.18653/v1/2022.acl-long.251} {Making
  transformers solve compositional tasks}.
\newblock In \emph{Proceedings of the 60th Annual Meeting of the Association
  for Computational Linguistics (Volume 1: Long Papers)}, pages 3591--3607,
  Dublin, Ireland. Association for Computational Linguistics.

\bibitem[{Peters et~al.(2018)Peters, Neumann, Zettlemoyer, and
  Yih}]{peters-etal-2018-dissecting}
Matthew Peters, Mark Neumann, Luke Zettlemoyer, and Wen-tau Yih. 2018.
\newblock \href {https://doi.org/10.18653/v1/D18-1179} {Dissecting contextual
  word embeddings: Architecture and representation}.
\newblock In \emph{Proceedings of the 2018 Conference on Empirical Methods in
  Natural Language Processing}, pages 1499--1509, Brussels, Belgium.
  Association for Computational Linguistics.

\bibitem[{Qiu et~al.(2021)Qiu, Shaw, Pasupat, Nowak, Linzen, Sha, and
  Toutanova}]{qiu2021improving}
Linlu Qiu, Peter Shaw, Panupong Pasupat, Pawe{\l}~Krzysztof Nowak, Tal Linzen,
  Fei Sha, and Kristina Toutanova. 2021.
\newblock Improving compositional generalization with latent structure and data
  augmentation.
\newblock \emph{arXiv preprint arXiv:2112.07610}.

\bibitem[{Raffel et~al.(2020)Raffel, Shazeer, Roberts, Lee, Narang, Matena,
  Zhou, Li, and Liu}]{raffel-etal-2020-t5}
Colin Raffel, Noam Shazeer, Adam Roberts, Katherine Lee, Sharan Narang, Michael
  Matena, Yanqi Zhou, Wei Li, and Peter~J. Liu. 2020.
\newblock \href {http://jmlr.org/papers/v21/20-074.html} {Exploring the limits
  of transfer learning with a unified text-to-text transformer}.
\newblock \emph{Journal of Machine Learning Research}, 21(140):1--67.

\bibitem[{Sennrich et~al.(2016)Sennrich, Haddow, and
  Birch}]{sennrich-etal-2016-controlling}
Rico Sennrich, Barry Haddow, and Alexandra Birch. 2016.
\newblock \href {https://doi.org/10.18653/v1/N16-1005} {Controlling politeness
  in neural machine translation via side constraints}.
\newblock In \emph{Proceedings of the 2016 Conference of the North {A}merican
  Chapter of the Association for Computational Linguistics: Human Language
  Technologies}, pages 35--40, San Diego, California. Association for
  Computational Linguistics.

\bibitem[{Stern et~al.(2017)Stern, Andreas, and
  Klein}]{stern-etal-2017-minimal}
Mitchell Stern, Jacob Andreas, and Dan Klein. 2017.
\newblock \href {https://doi.org/10.18653/v1/P17-1076} {A minimal span-based
  neural constituency parser}.
\newblock In \emph{Proceedings of the 55th Annual Meeting of the Association
  for Computational Linguistics (Volume 1: Long Papers)}, pages 818--827,
  Vancouver, Canada. Association for Computational Linguistics.

\bibitem[{Tenney et~al.(2019{\natexlab{a}})Tenney, Das, and
  Pavlick}]{tenney-etal-2019-bert}
Ian Tenney, Dipanjan Das, and Ellie Pavlick. 2019{\natexlab{a}}.
\newblock \href {https://doi.org/10.18653/v1/P19-1452} {{BERT} rediscovers the
  classical {NLP} pipeline}.
\newblock In \emph{Proceedings of the 57th Annual Meeting of the Association
  for Computational Linguistics}, pages 4593--4601, Florence, Italy.
  Association for Computational Linguistics.

\bibitem[{Tenney et~al.(2019{\natexlab{b}})Tenney, Xia, Chen, Wang, Poliak,
  McCoy, Kim, Durme, Bowman, Das, and
  Pavlick}]{DBLP:conf/iclr/TenneyXCWPMKDBD19}
Ian Tenney, Patrick Xia, Berlin Chen, Alex Wang, Adam Poliak, R.~Thomas McCoy,
  Najoung Kim, Benjamin~Van Durme, Samuel~R. Bowman, Dipanjan Das, and Ellie
  Pavlick. 2019{\natexlab{b}}.
\newblock \href {https://openreview.net/forum?id=SJzSgnRcKX} {What do you learn
  from context? probing for sentence structure in contextualized word
  representations}.
\newblock In \emph{7th International Conference on Learning Representations,
  {ICLR} 2019, New Orleans, LA, USA, May 6-9, 2019}. OpenReview.net.

\bibitem[{Weißenhorn et~al.(2022)Weißenhorn, Donatelli, and
  Koller}]{weissenhorn22starsem}
Pia Weißenhorn, Lucia Donatelli, and Alexander Koller. 2022.
\newblock \href {https://doi.org/10.18653/v1/2022.starsem-1.4} {Compositional
  generalization with a broad-coverage semantic parser}.
\newblock In \emph{Proceedings of the 11th Joint Conference on Lexical and
  Computational Semantics (*SEM)}.

\bibitem[{Yu et~al.(2019)Yu, Vu, and Kuhn}]{yu-etal-2019-learning}
Xiang Yu, Ngoc~Thang Vu, and Jonas Kuhn. 2019.
\newblock \href {https://doi.org/10.18653/v1/W19-4815} {Learning the {D}yck
  language with attention-based {S}eq2{S}eq models}.
\newblock In \emph{Proceedings of the 2019 ACL Workshop BlackboxNLP: Analyzing
  and Interpreting Neural Networks for NLP}, pages 138--146, Florence, Italy.
  Association for Computational Linguistics.

\bibitem[{Zheng and Lapata(2022)}]{zheng-lapata-2022-disentangled}
Hao Zheng and Mirella Lapata. 2022.
\newblock \href {https://doi.org/10.18653/v1/2022.acl-long.293} {Disentangled
  sequence to sequence learning for compositional generalization}.
\newblock In \emph{Proceedings of the 60th Annual Meeting of the Association
  for Computational Linguistics (Volume 1: Long Papers)}, pages 4256--4268,
  Dublin, Ireland. Association for Computational Linguistics.

\end{thebibliography}
\bibliographystyle{acl_natbib}

\appendix

\section{Training details} \label{appendix:sec:train}
\paragraph{Evaluation metrics.} 
We use sequence-level exact match accuracy as our evaluation metrics for all experiments. Thus a predicted sequence is correct only if each output token in it is correctly predicted.

\paragraph{Hyperparameters.}  
We used the following hyperparameter values in our experiments. For all experiments we reported, we use \textit{bart-base}\footnote{\url{https://huggingface.co/facebook/bart-base}} for BART model and \textit{t5-base}\footnote{\url{https://huggingface.co/t5-base}} for T5 model. We always use the Adam optimizer \cite{DBLP:journals/corr/KingmaB14} and gradient accumulation steps 8. Exact match accuracy is used as the validation metric.

We use the same hyperparameters setting for semantic parsing, syntactic parsing and POS tagging experiments. For BART, we use batch size 64 and learning rate 2e-4. For T5, we use batch size 32 and learning rate 5e-4. 

In probing experiments, we probe the encoder of BART. The hidden size of the MLP classifier is 1024 and the dropout is 0.1. We use batch size 64 and learning rate 1e-3 for the span prediction task and 1e-4 for the semantic role labeling task.

In the \cogsqa\ experiments, we adapt the question answering module\footnote{\url{https://huggingface.co/transformers/v4.5.1/model_doc/bart.html\#transformers.BartForQuestionAnswering}} of BART for BART-QA and BART-QA+struct. In the evaluation for such extractive models, we do not consider the capitalization of the determiners (e.g.\ \textit{The boy} is equivalent to \textit{the boy}).  We use batch size 64 and learning rate 2e-4 for these two models. For seq2seq models, most hyperparameters are the same as the ones in parsing tasks. The only difference is that we use learning rate 1e-4 for the T5 model.

\paragraph{Model selection.} \citet{csordas-etal-2021-devil} find that using an in-distribution development set can lead to inefficient model selection and they select their best model based on the accuracy on the generalization set. We follow \citet{zheng-lapata-2022-disentangled} by sampling a subset of the generalization set as an  out-of-distribution development set.  

\paragraph{In-distribution set performance.} The exact match accuracy is at least 99 for both the (in-distribution) development set and the (in-distribution) test set in all experiments for the parsing and tagging tasks. 

On the \cogsqabase\ dataset, all models (i.e.\ BART, T5 and BART-QA) achieve at least 99 accuracy on the in-distribution development and test sets. On the \cogsqacc\ dataset, we find T5 and BART-QA can achieve an accuracy of 100 on the in-distribution development and test sets across different randoms seeds. However, the performance of BART is not stable with regard to different random seeds. The mean accuracy averaged over 5 runs is 
$95\pm8.1$ 
for \textit{cc\_cp} and $73.8\pm27.8$ for \textit{rc\_pp}.

\paragraph{Other details.} Training takes 4 hours for BART with about 50 epochs and 4 hours for T5 with about 30 epochs. Inference on the  generalization set takes about 1 hour. All experiments are run on Tesla V100 GPU cards (32GB). The number of parameters is 140 million in BART and 220 million in T5. 

\paragraph{Results from other papers.}
\cite{kim-linzen-2020-cogs} provides two train sets: \textit{train} (24155 samples) and \textit{train100} (39500 samples). The \textit{train100} simply extends \textit{train} with 100 samples for each exposure example. For example, for the generalization type in Fig.\  \ref{fig:cogssamples} (a), \textit{train} set only contains 1 sentence with \textit{hedgehog} being the subject as the exposure example, but \textit{train100} contain 100 different sentences with \textit{hedgehog} being the subject. Since \textit{train100} does not introduce new structures, it is only used to help lexical generalization.

All semantic models in Table \ref{tab:selected_gentype_eval} are trained on the \textit{train} set, except for \cite{kim-linzen-2020-cogs,conklin-etal-2021-meta,weissenhorn22starsem}. We noticed that \cite{kim-linzen-2020-cogs,conklin-etal-2021-meta} get higher performance on \textit{train100} and thus report their number on \textit{train100}. Although the  number for \cite{weissenhorn22starsem} is based on \textit{train100}, their model actually performs well on structural generalization when trained on the \textit{train} set and using \textit{train100} only improves the performance on lexical generalization types. Thus their model still supports the point that structural generalization can be solved by structure-aware models.



\section{Dataset details} \label{appendix:sec:dataset}
We use COGS \cite{kim-linzen-2020-cogs} and variants of COGS (i.e. Syntax-COGS, POS-COGS and QA-COGS) as our datasets. We report dataset statistics for all our datasets in Table \ref{tab:dataset_statistics}. 

\begin{table*}[]
    \centering
    \begin{tabular}{lccccccc}
    \toprule
    Dataset & \# train & \# dev. & \# test & \# gen & Vocab. size & Train len. & Gen len. \\
    \midrule
    COGS & 24155 & 3000 & 3000 & 21000 & 871 & $22 / 153$ & $61 / 480$ \\
    \midrule
    Syntax-COGS & 24155 & 3000 & 3000 & 21000 & 759 & $22 / 129$ & $61 / 375$ \\
    \midrule
    POS-COGS & 24155 & 3000 & 3000 & 21000 & 753 & $22 / 21$ & $61 / 60$ \\
    \midrule
    QA-COGS-base & 54349 & 6834 & 6798 & 67989 & 793 & $44 / 19$ & $123 / 57$ \\
    \midrule
    \multirow{4}{*}{\textit{cc\_cp} (4 splits)} & 4000 & 1000 & 1000 & 2000 & 709 & $36 / 25$ & $35 / 18$ \\
     & 4000 & 1000 & 1000 & 2000 & 709 & $35 / 21$ & $36 / 25$ \\
     & 4000 & 1000 & 1000 & 2000 & 709 & $36 / 25$ & $30 / 19$ \\
     & 4000 & 1000 & 1000 & 2000 & 709 & $36 / 25$ & $34 / 21$ \\
     \midrule
    \multirow{4}{*}{\textit{rc\_pp} (4 splits)} & 4000 & 1000 & 1000 & 2000 & 594 & $19 / 5$ & $19 / 5$ \\
     & 4000 & 1000 & 1000 & 2000 & 594 & $19 / 5$ & $19 / 2$ \\
     & 4000 & 1000 & 1000 & 2000 & 594 & $19 / 5$ & $19 / 2$ \\
     & 4000 & 1000 & 1000 & 2000 & 594 & $19 / 5$ & $19 / 5$ \\
    \bottomrule
    \end{tabular}
    \caption{Statistics for all our datasets. \# denotes the number of instances in the dataset. Vocab.size denotes the size of vocabulary for the dataset, which consists of input tokens and output tokens. Train.len denotes the maximum length of the input tokens and output tokens in the train set. Gen.len denotes the maximum length in the generalization set.}
    \label{tab:dataset_statistics}
\end{table*}

\paragraph{Syntactic annotations.} To obtain syntactic annotations for Syntax-COGS, we use NLTK\footnote{\url{https://www.nltk.org/}} to parse each sentence in COGS with the context-free grammar that was used to generate COGS. In our experiments, we find this parsing process yields a unique tree for each sentence in COGS. The original grammar contains rules such as \lform{NP$\to$NP\_animate\_dobj\_noPP}. We replace such fine-grained nonterminals (e.g.\ \lform{NP\_animate\_dobj\_noPP}) with general nonterminals (e.g.\ \lform{NP}). This results in duplicate patterns (e.g.\ \lform{NP$\to$NP}) and we further remove such patterns from the output tree. 

\section{Semantic role labeling}  \label{appendix:sec:srl}
We give more details about semantic role labeling task described in  \cref{sec:probing-howto} here. In contrast to the semantic parsing task, where the output is a sequence encoding the meaning representation, the goal of this task is to predict the semantic role graph of a sentence.

\begin{figure}
    \centering
    \includegraphics[scale=0.5]{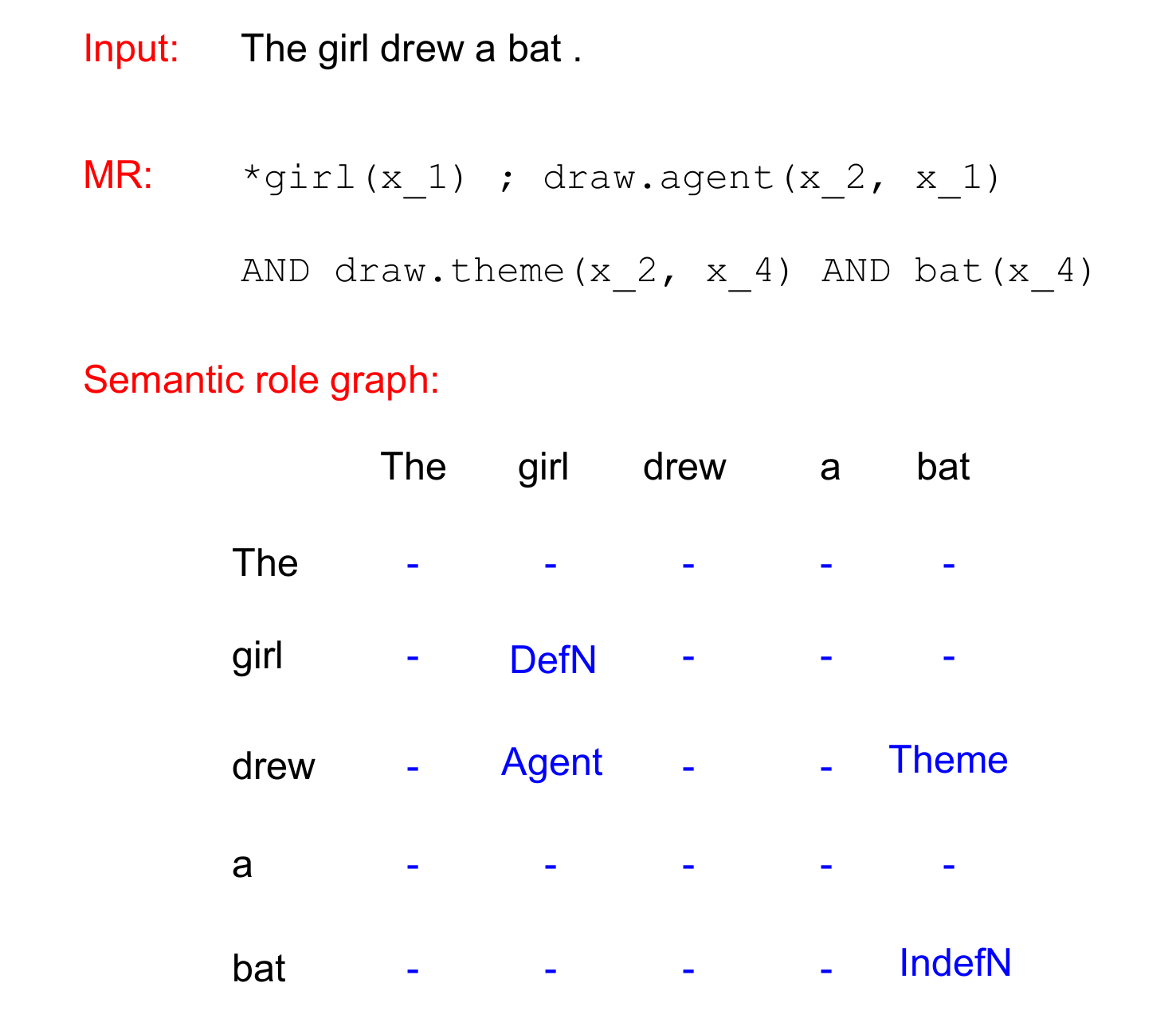}
    \caption{An example to show how to transform a meaning representation to a semantic role graph.}
    \label{fig:srl_example}
\end{figure}

An example of the semantic role graph is shown in \cref{fig:srl_example}. The symbol \texttt{-} denotes that the column word is not an argument of the row word; we capture this with the special class  \textit{None} in the data. 

We align tokens in the sentence and predicate symbols in the meaning representation based on the variable names, which specify positions in the sentence (e.g.\ $x_1$ corresponds to the second token in the string). This allows us to project the predicate-argument relations in the meaning representation to relations between the tokens. For a predicate verb, we connect an edge to each of its arguments (i.e.\ \textit{drew} has an \texttt{Agent} edge to \textit{girl.}) in the meaning representation. 

The COGS grammar also contains prepositional phrases (e.g.\ \textit{a bowl on the table}). To represent this modification relation, we connect an \texttt{Nmod} edge from the modified noun to the modifier noun (e.g.\ \textit{bowl} has an \texttt{Nmod} edge to \textit{table}). 

For common nouns, we connect a \texttt{DefN} edge to itself to denote it has a definite determiner (e.g.\ \textit{girl} has a \texttt{DefN} edge to itself) and a \texttt{IndefN} to denote it has an indefinite determiner (e.g.\ \textit{bat} has an \texttt{IndefN} edge to itself).

\section{QA-COGS}  \label{appendix:sec:qa}
\subsection{QA-COGS-base}
To create QA-COGS-base, we first obtain the frame for each predicate in a sentence from its gold meaning representation. We define the frame of a predicate as the combination of argument types it takes. Possible frames in our dataset and corresponding examples are shown in Table \ref{tab:qacogssamples}. We generate questions for all predicate-argument pairs in a sentence. Thus a sentence with two predicates both of which takes two arguments will result in 4 question-answer pairs. 

\begin{table*}[htb!]
    \small
    \centering
    \setlength{\tabcolsep}{2pt}
    \begin{tabularx}{\linewidth}{l|X|X|X} \toprule
         \textbf{Predicate Frame} & \textbf{Context} & \textbf{Question} & \textbf{Answer} \\ 
         \midrule
         {\genclass{agent}} & {The captain ate}  & {Who ate ?} & {the captain}\\
         \midrule
         {\genclass{theme}} & {The donut was known}  & {What was known ?} & {the donut}\\
         \midrule
         \multirow{2}{*}{\genclass{agent\_theme}} & \multirow{2}{3cm}{Emma ate the ring beside a bed}  & {What did Emma eat. ?} & {the ring beside a bed} \\
         & & {Who ate the ring beside a bed ?} & {Emma} \\   
         \midrule
         \multirow{3}{*}{\genclass{agent\_theme\_recipient}} & \multirow{3}{3cm}{Amelia gave Emma a strawberry}  & {Who gave a strawberry to Emma ?} & {Amelia} \\
         & & {What did Amelia give to Emma ?} & {a strawberry} \\
         & & {Who did Amelia give a strawberry to ?} & {Emma} \\
         \midrule
         \multirow{2}{*}{\genclass{theme\_recipient}} & \multirow{2}{3cm}{A rose was mailed to Isabella}  & {Who was a rose mailed to ?} & {Isabella}\\
         & & {What was mailed to Isabella ?} & {a rose} \\
         \midrule
         \multirow{2}{*}{\genclass{agent\_ccomp}} & \multirow{2}{3cm}{Liam meant that Sophia  rolled a teacher on a seat}  & {What did Liam mean ?} & {that Sophia rolled a teacher on a seat} \\
         & & {Who meant that Sophia rolled a teacher on a seat ?} & {Liam} \\
         \midrule
         \multirow{2}{*}{\genclass{agent\_xcomp}} & \multirow{2}{*}{Emma hoped to run}  & { Who hoped to run ?} & {Emma}\\
         & & {What did Emma hope to do ?} & {run} \\
         \bottomrule
    \end{tabularx}
    \caption{
        All possible predicate frames and corresponding question-answer examples for the QA-COGS-base dataset.
    }\label{tab:qacogssamples}
\end{table*}

\subsection{QA-COGS-disamb}
We adapt the original context-free grammar from which the COGS training set was generated and make some changes to it to generate QA-COGS-disamb. We refer readers to Appendix A and B in \citet{kim-linzen-2020-cogs} for more details of the original grammar.

For cc\_cp, we introduce the coordination structure and present tense into the grammar. We also simplify the grammar by removing the grammar rule for passive verbs (e.g.\ \textit{eaten}) and subject control verbs that take infinitival arguments (e.g.\ \textit{try}). We do this to avoid such verbs resulting in ambiguous sentences (e.g. \textit{Oliver said that Noah is helped and painted}). The grammar enforces that the tense of the complement clause must be different from the one in the main clause to avoid ambiguity (that is, if the verb in the main clause is in past tense, then the verb in the subordinate clause must be in present tense). We extend the verb vocabulary with their present tenses and use the same noun vocabulary as COGS. 

For rc\_pp, we add relative clauses and plural nouns to the grammar. We also simplify the grammar by removing the grammar rule for verb phrases that do not have a common noun as object, e.g.\ verbs taking CP arguments (e.g.\ \textit{say}) and unaccusative verbs (e.g. \textit{sleep}). The grammar enforces that the head noun of the NP and the head noun of the PP differ in number (that is, if the NP is singular, then the PP must be plural). In original COGS grammar, the vocabularies for nouns in NPs and PPs are separate (e.g.\ \textit{a cake on the table}, \textit{table} can only appear after \textit{on}). We change this by using the same noun vocabulary for both. We also extend the noun vocabulary with their plural forms and extend the verb vocabulary with \textit{were}. 

\section{Detailed results}
We report detailed results for our best models in Table \ref{tab:detailed_results}. We report averaged accuracy and the standard deviation over 5 runs. \textit{BART+mtl} and \textit{BART+mask} denotes the model we used in section \ref{subsec:seq2seq:semantic_gen_with_syntax}. 

\begin{table*}[]
    \centering
    \resizebox{\linewidth}{!}{
    \begin{tabular}{ll|ccc|c|c}
    \toprule
    & & \multicolumn{3}{c|}{\structg} & \lexg &   \\
    Dataset & Model & Obj to Subj PP & CP recursion & PP recursion & all 18 other types & overall \\
    \midrule
    \multirow{5}{*}{COGS} & BART & $0.0 \pm 0.0$ & $0.5 \pm 0.2$ & $11.8 \pm 1.5$ & $91.1 \pm 0.4$ & $78.6 \pm 0.3$ \\
    & BART+syn & $0.0 \pm 0.0$ & $5.3 \pm 0.8$ & $7.7 \pm 0.3$ & $92.8 \pm 0.5$ & $80.1 \pm 0.5$ \\
    & BART+mtl & $0.0 \pm 0.0$ & $0.3 \pm 0.2$ & $10.9 \pm 1.9$ & $92.1 \pm 0.3$ & $79.5 \pm 0.3$ \\
    & BART+mask & $0.0 \pm 0.0$ & $0.1 \pm 0.1$ & $4.9 \pm 2.9$ & $84.0 \pm 2.8$ & $72.2 \pm 2.5$ \\
    & T5 & $0.0 \pm 0.0$ & $0.0 \pm 0.0$ & $8.6 \pm 2.2$ & $96.9 \pm 0.3$ & $83.5 \pm 0.2$ \\
    \midrule
    \multirow{2}{*}{Syntax-COGS} & BART & $0.0 \pm 0.0$ & $9.1 \pm 1.5$ & $22.3 \pm 1.1$ & $99.5 \pm 0.1$ & $86.8 \pm 0.1$ \\
    & T5 & $4.7 \pm 8.7$ & $7.2 \pm 1.3$ & $9.0 \pm 4.0$ & $99.4 \pm 0.5$ & $86.2 \pm 0.3$ \\
    \midrule
    \multirow{2}{*}{POS-COGS} & BART & $0.0 \pm 0.0$ & $5.8 \pm 5.2$ & $19.1 \pm 10.2$ & $97.9 \pm 1.0$ & $85.1 \pm 1.1$ \\
    & T5 & $0.0 \pm 0.0$ & $4.2 \pm 2.5$ & $3.9 \pm 4.1$ & $98.1 \pm 1.1$ & $84.5 \pm 0.9$ \\
    \midrule
    \multirow{3}{*}{QA-COGS-base} & BART & $98.9 \pm 0.7$ & $58.8 \pm 4.3$ & $69.1 \pm 1.0$ & $95.3 \pm 0.3$ & $85.7 \pm 0.7$ \\
    & T5 & $100.0 \pm 0.0$ & $94.7 \pm 2.3$ & $96.9 \pm 0.6$ & $100.0 \pm 0.0$ & $98.6 \pm 0.5$ \\
    & BART-QA & $100.0 \pm 0.0$ & $97.6 \pm 0.9$ & $99.6 \pm 1.0$ & $100.0 \pm 0.0$ & $99.2 \pm 0.4$ \\
    \midrule
    \multirow{4}{*}{\textit{cc\_cp}} & BART & - & - & - & - & $36.5 \pm 22.0$ \\
    & T5 & - & - & - & - & $15.6 \pm 1.7$ \\
    & BART-QA  & - & - & - & - & $5.6 \pm 10.0$ \\
    & BART-QA+struct  & - & - & - & - & $100.0 \pm 0.0$ \\ 
    \midrule
    \multirow{4}{*}{\textit{rc\_pp}} & BART & - & - & - & - & $13.7 \pm 5.1$ \\
    & T5 & - & - & - & - & $21.9 \pm 2.0$ \\
    & BART-QA  & - & - & - & - & $0.0 \pm 0.0$ \\
    & BART-QA+struct  & - & - & - & - & $100.0 \pm 0.0$ \\ 
    \bottomrule
    \end{tabular}
    }
    \caption{Detailed results for our models across COGS, Syntax-COGS, POS-COGS and QA-COGS.}
    \label{tab:detailed_results}
\end{table*}

%

\end{document}